\documentclass{Interspeech}

\interspeechcameraready

\title{The Interspeech 2025 Speech Accessibility Project Challenge}

\author[affiliation={1}]{Xiuwen}{Zheng}
\author[affiliation={1}]{Bornali}{Phukon}
\author[affiliation={8}]{Jonghwan}{Na}
\author[affiliation={2}]{Ed}{Cutrell}
\author[affiliation={3}]{Kyu}{Han}
\author[affiliation={1}]{Mark}{Hasegawa-Johnson}
\author[affiliation={4}]{Pan-Pan}{Jiang}
\author[affiliation={2}]{Aadhrik}{Kuila}
\author[affiliation={5}]{Colin}{Lea}
\author[affiliation={4}]{Bob}{MacDonald}
\author[affiliation={5}]{Gautam}{Mantena}
\author[affiliation={3}]{Venkatesh}{Ravichandran}
\author[affiliation={6}]{Leda}{Sari}
\author[affiliation={7}]{Katrin}{Tomanek}
\author[affiliation={9}]{Chang D.}{Yoo}
\author[affiliation={1}]{Chris}{Zwilling}

\affiliation{}{UIUC}{}
\affiliation{}{Microsoft}{}
\affiliation{}{Amazon}{}
\affiliation{}{Google}{}
\affiliation{}{Apple}{}
\affiliation{}{Meta}{USA}
\affiliation{}{CDLI}{UK}
\affiliation{}{Inha University}{}
\affiliation{}{KAIST}{KR}
\email{\{xiuwenz2, bornalip, jhasegaw, zwillin1\}@illinois.edu, \{cutrell, aadhrikkuila\}@microsoft.com, \{kyujhan, veravic\}@amazon.com, \{pjiang, bmacdonald\}@google.com, \{colin.lea, gmantena\}@apple.com, ledassari@gmail.com, katrin.tomanek@gmail.com, jhna@dsp.inha.ac.kr, cd\_yoo@kaist.ac.kr}

\usepackage{comment}  
\usepackage[table,xcdraw]{xcolor}
\usepackage{siunitx}
\usepackage{amsmath}
\usepackage{xurl}
\usepackage{hyperref}

\begin{document}

\maketitle

\begin{abstract}
While the last decade has witnessed significant advancements in Automatic Speech Recognition (ASR) systems, performance of these systems for individuals with speech disabilities remains inadequate, partly due to limited public training data. To bridge this gap, the 2025 Interspeech Speech Accessibility Project (SAP) Challenge was launched, utilizing over 400 hours of SAP data collected and transcribed from more than 500 individuals with diverse speech disabilities. Hosted on EvalAI and leveraging the remote evaluation pipeline, the SAP Challenge evaluates submissions based on Word Error Rate and Semantic Score. Consequently, 12 out of 22 valid teams outperformed the whisper-large-v2 baseline in terms of WER, while 17 teams surpassed the baseline on SemScore. Notably, the top team achieved the lowest WER of 8.11\%, and the highest SemScore of 88.44\% at the same time, setting new benchmarks for future ASR systems in recognizing impaired speech.
\end{abstract}

\section{Introduction}
\label{sec:introduction}

Automatic Speech Recognition (ASR) has witnessed remarkable advancements in recent years, primarily driven by the development of deep neural networks (DNN) and the explosive growth of training data. End-to-end ASR systems, in particular, leveraging self-supervised learning (SSL) or large-scale weakly-supervised learning techniques, have become the standard paradigm which achieved significant improvements in recognition accuracy and overall performance. These progressions have enabled ASR technologies to become increasingly integrated into various applications.

However, a notable performance gap remains when recognizing speech from individuals with disabilities compared to individuals without disabilities. As reported in~\cite{hasegawa2024community}, between 2008 and 2023, the lowest error rates for dysarthric ASR decreased by a factor of three, while for individuals without dysarthria, the error rates decreased by a factor of five. One primary cause of this disparity is the limited availability of large-scale, diverse data for the standardized and repeatable training and testing of ASR models against dysarthric speech: The largest publicly available English corpus meeting these criteria from 2008 to 2023 was the UA-Speech~\cite{kim08c_interspeech} corpus, which includes 22 hr of speech by 16 speakers. To address the scarcity of specialized data that prevents ASR systems from effectively generalizing and adapting to impaired speech, the Speech Accessibility Project (SAP~\cite{hasegawa2024community}) is systematically collecting, transcribing, and distributing U.S., Canadian, and Puerto Rican English speech from individuals with speech disorders. This effort aims to prompt research and development in impaired speech recognition, thereby enhancing ASR technology for people with diverse speech patterns and disabilities.

In alignment with these objectives, we launched the Interspeech 2025 Speech Accessibility Project (SAP) Challenge, aimed at rapidly advancing the state of the art in dysarthric speech recognition. To the best of our knowledge, this is the first impaired speech recognition challenge utilizing a large-scale speaker-independent 
corpus.
The challenge utilizes the partial SAP dataset released on April 30, 2024, containing over 400 hours of speech data collected from more than 500 individuals with disabilities. For the remainder of this paper, we will refer to this dataset as SAP-240430. We host the SAP challenge\footnote{\url{https://eval.ai/web/challenges/challenge-page/2362}} through EvalAI~\cite{yadav2019evalai}. To protect data privacy, we utilize EvalAI's remote evaluation pipeline, ensuring that evaluations are conducted on a private server provided by The National Center for Supercomputing Applications (NCSA), which securely stores the test data. Competition model submissions are assessed based on two primary metrics: Word Error Rate (WER) and Semantic Score (SemScore). WER serves as a traditional measure of ASR accuracy by quantifying the discrepancies between the recognized words and the ground truth transcripts. SemScore, on the other hand, evaluates the semantic fidelity of the transcriptions, assessing the degree to which the transcription preserves the intended meaning and context. We divided the test data into two subsets: Test1 and Test2. Results from Test1 are used to benchmark participants on a public leaderboard, while results from Test2 remain hidden on a private leaderboard until the competition concludes. Final team rankings are based on the outcomes from the private leaderboard. As a result, the SAP Challenge received valid results from 22 teams, 12 of which surpassed the baseline (whisper-large-v2) on WER, and 17 showed better SemScore. The leading team attained the lowest WER of 8.11\%, as well as the highest Semscore of 88.44\%, establishing new benchmarks for future ASR systems in the domain of impaired speech recognition.

\section{Data, Metrics, and Baseline}
\subsection{Data}
\label{subsec:data}
In this section, we provide a comprehensive overview of the SAP-240430 dataset utilized in the SAP Challenge. Comprising approximately 415 hours of impaired speech, the dataset includes contributions from 524 participants diagnosed with one of the following five etiologies: Parkinson's Disease (PD), Down Syndrome (DS), amyotrophic lateral sclerosis (ALS), cerebral palsy (CP), or stroke. Contributors to the corpus were recruited by the University of Illinois, in collaboration with patient advocacy organizations; all procedures were approved by the Institutional Review Board at the University of Illinois. Speech for this competition was primarily recorded during the period April through December 2023, when the Speech Accessibility Project was only recruiting participants with PD, therefore the dataset for this competition was dominated by speech from people with Parkinsons Disease: 73.4\% of the training corpus, and 75.9\% of the test corpus, by speech duration, were recorded by people with PD.  The most common type of dysarthria for people with PD is hypokinetic dysarthria~\cite{darley1969differential}, which is characterized by constant low energy levels, flat pitch, articulatory reductions, and voicing that is often breathy or even whispered~\cite{solomon1993speech,ramig1995comparison}.  Some speakers, however, exhibited speech patterns that differed markedly from the norm, including seven speakers who produced extremely long sequences of stutter events, and others who exhibited spasticity, extremely rapid speech, and extremely slow speech.
The dataset is partitioned into Training, Development, and Test sets following a 70:10:20 ratio, ensuring that each split contains unique speakers with no overlaps on the speaker-level. Participants were selected for dev and test sets with equal probability regardless of etiology. Furthermore, the Test set is subdivided into two equal parts, Test1 and Test2. During inference, we employ the ``unshared" test subsets, which excludes any utterances whose text also appears in the training data. As outlined in section \ref{sec:introduction}, the evaluation results from the Test1 ``unshared" subset are publicly available for benchmarking purpose, while the final rankings of participants are determined based on the results from the Test2 ``unshared" subset.

\begin{table}[t!]
  \begin{centering}
  \caption{Statics about the (processed) SAP-240430 dataset for each split, including the number of speakers (\# of speakers), number of utterances (\# of utterances), and the duration of speech data in hours (Duration (\SI{}{\hour})).} \label{table:sap0430info}
  \scalebox{0.93}{
    \begin{tabular}{l||c|c|c}
    \toprule
    \textbf{Split} & \textbf{\# of speakers} & \textbf{\# of utterances} & \textbf{Duration (\SI{}{\hour})}\\
     \midrule
     Train & 369 & 131,420 & 290.35 \\
     Dev & 55 & 19,275 & 43.56 \\
     Test1 & 50 & 18,397 & 42.16 \\
     \rowcolor[HTML]{EFEFEF}
     \hspace{0.35em}unshared & 50 & 7,601 & 25.17 \\
     Test2 & 50 & 17,752 & 38.77 \\
     \rowcolor[HTML]{EFEFEF}
     \hspace{0.35em}unshared & 50 & 8,043 & 23.39 \\
    \bottomrule
    \end{tabular}}
\end{centering}
\end{table}

Participants are provided with the unprocessed and processed Train and Dev sets, and the corresponding pre-processing pipeline\footnote{\label{sharedfootnote}\url{https://github.com/xiuwenz2/SAPC-template}}. The speech data was down-sampled to a standardized sampling rate of \SI{16}{\kilo\hertz}, and transcriptions were normalized according to the following rules:
\begin{itemize}[labelindent=1em, leftmargin=1em]
\item Words enclosed in square brackets were removed, including spontaneous speech prompts, 
and speech-language pathologist (SLP) comments, where no associated audio exists.
\item The notation 
\{g:$w$\}, used to denote annotator uncertainty (``guess'') about the word $w$, was replaced by an instance of the word $w$ without markup.  Symbols denoting unknown words were all 
replaced with ``UNK" to ensure consistent treatment of ambiguous content.
\item To support the development of ASR systems capable of handling disfluencies effectively, two transcript versions were provided: one including disfluencies and self-corrections (reparanda)~\cite{levelt1983monitoring,howell1991use}, each enclosed in parentheses, and another with disfluencies and reparanda removed.
\item Basic text normalization was performed using the NeMo Text Normalization Toolkit~\cite{zhang21ja_interspeech}, converting elements such as numeric digits, special punctuation marks and abbreviations into their corresponding spoken equivalents. Subsequently, a comprehensive manual review was conducted to address normalization errors and insert spaces between the letters of abbreviations.
\item To ensure consistency in the final transcripts, all punctuation was eliminated except for apostrophes within words, and the text was transformed to uppercase.
\end{itemize}
The detailed information about each data subset within SAP-240430 (after data pre-processing) is presented in Table \ref{table:sap0430info}.

\subsection{Metrics}

The SAP Challenge employs word error rate and semantic score as evaluation metrics. Given that ASR models may handle disfluencies differently, two versions of reference transcripts—one including disfluencies and the other excluding them—were prepared, as described in Section \ref{subsec:data}. During evaluation, the transcript version that optimizes the metric for each test utterance (i.e., lower WER or higher SemScore) is selected dynamically. The top-performing team for each metric is recognized, establishing benchmarks to guide future research. To promote transparency and reproducibility, the evaluation scripts\textsuperscript{\ref{sharedfootnote}} are publicly available.

Word Error Rate is a standard metric for evaluating ASR system performance. It quantifies transcription accuracy by calculating the normalized string edit distance between the hypothesis generated by the ASR system and the ground truth reference. 
To eliminate potential bias from unusually high WER values, we limited the maximum WER to 100\% for each utterance. Some downstream applications use ASR to capture the details of an utterance (including disfluencies and reparanda), while others seek to capture only the intended utterance, therefore the WER for each utterance was calculated by comparing the hypothesis transcript to two different reference transcripts (with and without disfluencies and reparanda), and choosing the smaller of the two.
Consequently, for a given test set consisting of $M$ utterances, the formula used in this challenge to calculate WER is as follows:
\begin{align}
\text{WER} &= \frac{\sum_{i=1}^M \text{WER}_i \cdot N_i^{*}}{\sum_{i=1}^M N_i^{*}} \label{eq:uttwer} \\
\text{WER}_i &= \min\left( 1, \min_{j \in \{0,1\}} \left( \frac{S_{i,j} + D_{i,j} + I_{i,j}}{N_{i,j}} \right) \right)
\end{align}

where $S_{i,j}$, $D_{i,j}$, $I_{i,j}$, and $N_{i,j}$ represent utterance-level counts of substitutions, deletions, insertions, and words in the $i^{\text{th}}$ reference transcript with ($j=0$) or without ($j=1$) disfluencies, respectively, and $N_i^*$ is denominator of the minimizer of Eq.~(\ref{eq:uttwer}).

Semantic Score redefines ASR evaluation by shifting the focus from word-level accuracy to assessing how effectively transcriptions preserve the utterance-level intended meaning and contextual coherence. The SAP Challenge employs a comprehensive version of SemScore introduced in~\cite{Phukon2024}.  This SemScore was designed to capture the degree to which an ASR transcript is understandable by human readers. Six human readers each compared ASR hypothesis transcripts to reference transcripts, and used a Likert scale to label the degree to which the speaker's meaning can be understood from the ASR hypothesis. 
These human ratings were then used to determine linear regression weights that combine three objective scores: the MENLI logical entailment score~\cite{chen2023menli}, which employs natural language inference (NLI) to evaluate whether the logical content of the reference transcript is maintained in the ASR hypothesis; the BertScore F1 semantic similarity score~\cite{zhang2019bertscore}, which uses contextual embedding to measure the semantic similarity between the hypothesis and reference transcript; and the Soundex Coding phonetic distance\footnote{\url{https://ics.uci.edu/~dan/genealogy/Miller/javascrp/soundex.htm}\label{sharedfootnote2}}, which quantifies the degree of phonetic similarity between the hypothesis and the reference transcript, thereby assessing how accurately original pronunciation is preserved.
\begin{equation}
\text{SemScore} = \alpha\cdot\text{Score}_\text{NLI}+\beta\cdot\text{Score}_\text{BERT}+\gamma\cdot\text{Score}_\text{Soundex},
\end{equation}
where $\text{Score}_\text{NLI}$ is computed using a fine-tuned RoBERTa-large model~\cite{liu2019roberta}, while $\text{Score}_\text{BERT}$ is obtained using the official codebase\footnote{\nolinkurl{https://github.com/Tiiiger/bert_score}}. The $\text{Score}_\text{Soundex}$ is calculated using Soundex with Jaro-Winkler similarity\textsuperscript{\ref{sharedfootnote2}}. Hyperparameters $\alpha, \beta, \gamma$ are empirically determined\footnote{$\alpha=0.40, \beta=0.28, \gamma=0.32$} via linear regression with 5-fold cross-validation using human-rated ASR hypothesis-reference pairs.

\subsection{Baseline System}

\begin{table}[t!]
\centering
  \caption{Baseline model results on SAP-240430 Test1/Test2 unshared subsets. *Wav2vec-based models are all fine-tuned on LibriSpeech-960h.}\label{table:baseline_results}
  \scalebox{0.95}{
    \begin{tabular}{l|l||c|c}
    \toprule
    \multicolumn{2}{l||}{\textbf{Model}} & \textbf{WER \% $\downarrow$} & \textbf{SemScore \% $\uparrow$}\\
     \midrule
     whisper & base & 27.17/31.08 & 68.16/62.41 \\
     \rowcolor[HTML]{EFEFEF}
      & base.en & 20.87/24.49 & 73.68/68.25 \\
      & large.v2 & \textbf{14.97/17.82} & \textbf{82.26}/75.85 \\
     \rowcolor[HTML]{EFEFEF}
      & large.v3 & 21.06/22.81 & 81.82/75.89 \\
      & large.v3.turbo & 22.48/22.40 & 81.48/\textbf{76.04} \\ \midrule
      wav2vec\textsuperscript{*} & base & 33.90/39.16 & 64.03/57.49 \\
      \rowcolor[HTML]{EFEFEF}
      & large & 27.82/33.08 & 69.00/61.70 \\
      & large.robust & 22.80/26.55 & 74.91/68.77 \\
    \bottomrule
    \end{tabular}
    }
\end{table}

In order to ensure that the competition baseline is easily accessible to all participants, open-source ASR systems with publicly available model parameters were tested without any further finetuning, and the system with the lowest WER on the Test1 and Test2 unshared subsets was chosen as competition baseline. As shown in Table~\ref{table:baseline_results}, the official whisper-large-v2\footnote{\url{https://github.com/openai/whisper}} model, featuring 1.55 billion parameters, achieved the best WER, with 14.97\% on Test1 and 17.82\% Test2. Consequently, it was chosen as the challenge baseline. Part of OpenAI's Whisper~\cite{radford2023robust} family, the model leverages a transformer-based encoder-decoder architecture and scales weakly supervised pre-training to an extensive 680,000 hours of audio, with 117,000 hours covering 96 languages beyond English.



\begin{figure}[t!]
    \scalebox{1.0}{
    \includegraphics[trim=5mm 0mm 0mm 0mm, clip, width=\linewidth]{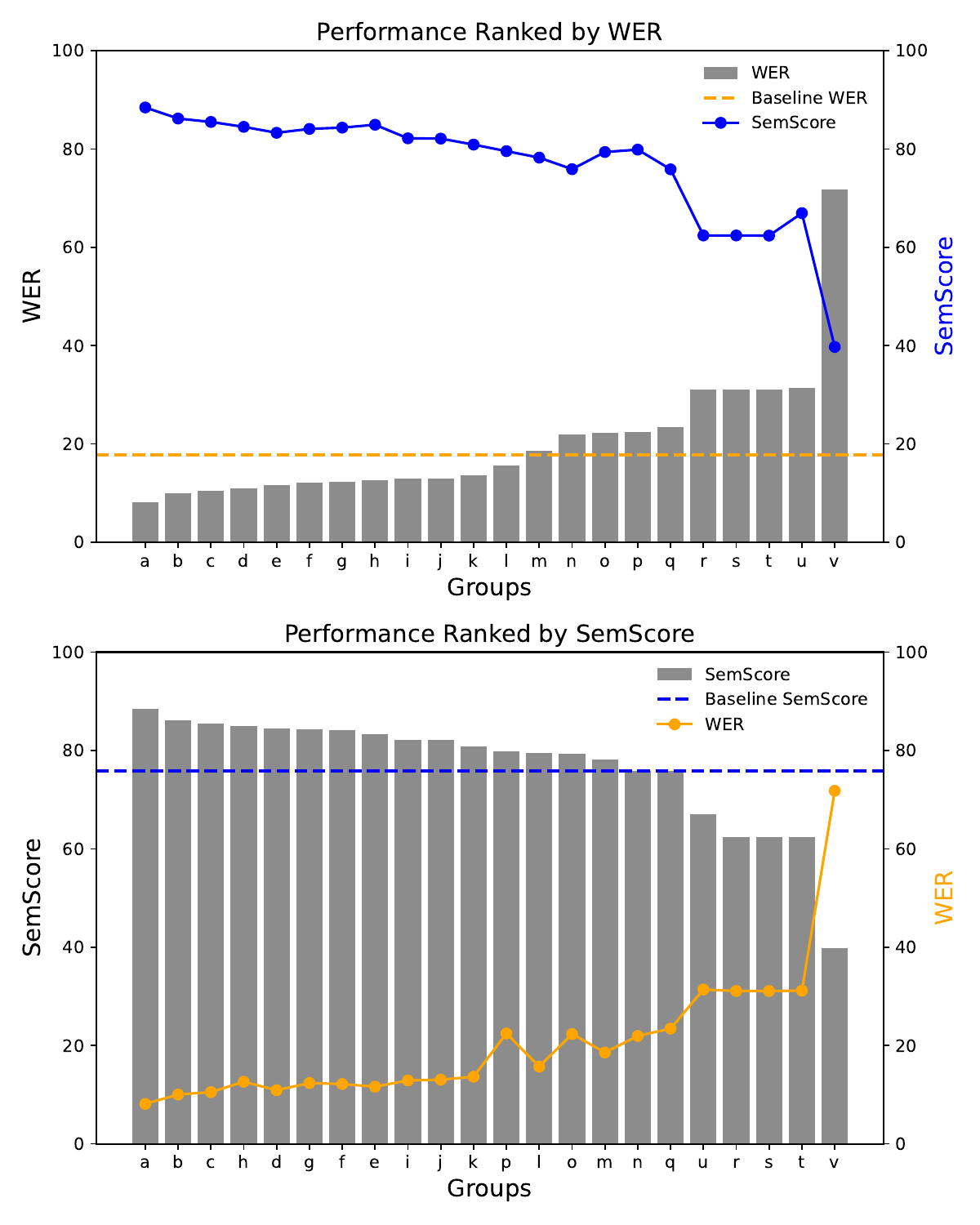}}
    \caption{Ranked performance of all participating teams by WER \% (top) and SemScore \% (bottom), with baseline results. Note that for a single team, the best WER and the best SemScore may come from different submissions.}
    \label{fig:overall_results}
\end{figure}

\begin{table}[t!]
\centering
  \caption{Performance (WER \%, SemScore \%) of top 5 teams and the baseline, on SAP-240430 Test2/unshared subset}\label{table:performance_analysis}
  \scalebox{1.0}{
    \begin{tabular}{c|c||c|c}
    \toprule
    \textbf{Team} & \textbf{WER \% $\downarrow$} & \textbf{Team} & \textbf{SemScore \% $\uparrow$}\\
     \midrule
     a & 8.11 & a & 88.44 \\
     \rowcolor[HTML]{EFEFEF}
     b & 10.03 & b & 86.19 \\
     c & 10.51 & c & 85.50 \\
     \rowcolor[HTML]{EFEFEF}
     d & 10.90 & h & 84.93 \\
     e & 11.62	& d & 84.50 \\
     \midrule
     * & 17.82 & * & 75.85 \\
    \bottomrule
    \end{tabular}
    }
    \vspace{-0.5em}
\end{table}

\section{Results and Discussion}

Around 40 parties joined the SAP Challenge by signing the Data User Agreement (DUA)\footnote{https://speechaccessibilityproject.beckman.illinois.edu/conduct-research-through-the-project}. Among these participants, 22 teams progressed to the submission stage, benchmarking their performance on the SAP-240430 Test1/unshared split, with results displayed on the public leaderboard. Figure~\ref{fig:overall_results} illustrates the ranked final outcomes of all participating teams on the Test2/unshared split, which was held until the conclusion of the challenge. To protect the privacy of the participants, the teams' original names have been replaced with the letters a–v throughout the paper. Notably, 12 out of 22 teams outperformed the baseline system on WER, while 17 teams achieved a higher SemScore.

As detailed in Table~\ref{table:performance_analysis}, the top 5 teams achieved comparable results, each demonstrating strong performance. The leading team achieved a WER of 8.11\%, and a SemScore of 88.44\%, which represent relative improvements of 54.49\% and 16.60\% over the baseline, respectively.

We conducted additional analyses on the 29 top-performing systems from each team. In cases where a team’s best WER and best SemScore came from different systems, both systems were included. SemScore is computed as a linear combination of logical entailment, semantic similarity, and phonetic distance, therefore there is no {\em a priori} necessity for its rankings to closely match the system rankings produced by WER. Nevertheless, we find that WER and SemScore exhibit a highly significant negative correlation, with a Pearson correlation coefficient of $\rho=-0.9649$. In most cases, rank order differences between the two scoring metrics were observed when systems had very similar scores. Among the 22 participating teams, 7 teams achieved their best WER and SemScore through two separate submissions. On average, the team-wise difference between the paired submissions is 0.92\% (std: 1.47) for WER and 1.12\% (std: 1.24) for SemScore. We analyzed the system pair that exhibited the largest performance differences across our two metrics, and the following example illustrates that a lower WER does not necessarily correspond to a higher SemScore. Given the reference sentence `how do you spell exercise,' the two ASR systems produced the hypotheses `how do you feel exercise' and `how to spell exercise.' 
The first hypothesis retains more of the reference words but less of the reference meaning (WER=20\%, SemScore=41.84\%); the second hypothesis retains fewer reference words but more reference meaning (WER=40\%, SemScore=90.36\%).

The baseline Whisper system tends to omit disfluent segments in transcription. In terms of WER, 3.48\% of the Test2 hypotheses favor references containing disfluencies (Type 1), while 11.86\% prefer references without disfluencies (Type 2). The remaining 84.66\% shows no preference, either because the references contain no disfluency or the WER is identical for both references types. As for SemScore, 12.74\% prefer reference Type 1, whereas 21.6\% choose Type 2. Among the 29 selected systems, only three exhibit a preference for transcribing disfluencies both in terms of WER and SemScore, two of which are from team \textit{e}. This tendency may result from a combination of the foundation model’s architecture and the type of references used for training and fine-tuning.

For etiology-specific analysis, we focus on individuals with PD and ALS due to the limited number of speakers in the Test 2 split for DS, CP, and Stroke. The baseline model achieves comparable WERs of 17.09\% for PD and 16.88\% for ALS while the top five models ranked by WER reduce these to an average of 10.06\% and 7.36\%, respectively. For SemScore, the baseline model achieves 77.99\% for PD and 74.08\% for ALS, with the top five models improving these to an average of 86.91\% and 88.11\%. Despite PD being the dominant etiology in the dataset, the relative improvement for ALS is larger, possibly due to lower variability among ALS speakers in the Test2 split. Specifically, the baseline model's average speaker-level WERs are 18.47\% (std:0.1933) for PD and and 16.29\% (std:0.1032) for ALS, and the best model from team \textit{a} reduces these to 9.61\% (std:0.1401) and 5.05\% (std:0.0330).

To gain deeper insights, we analyze the architectures, training strategies and key techniques adopted by the top-performing teams. As presented in Table~\ref{table:training_strategies}, all proposed ASR systems build upon existing large, publicly available foundation ASR models, namely parakeet from NVIDIA\footnote{https://huggingface.co/collections/nvidia/parakeet-659711f49d1469e51546e021} and whisper from OpenAI. By fine-tuning these models on the SAP data, they achieve substantial improvements in ASR performance and robust generalization for unseen speakers with dysarthria. In addition, team \textit{a} and \textit{b} both adopt audio segmentation strategies by dividing long audio files into shorter clips. To improve model robustness, Team \textit{a} employs model‐merging that combines multiple checkpoints, while Team \textit{c} addresses the hallucination problem using WhisperX preprocessing pipeline and rule-base postprocessing. Team \textit{d} boosts overall ASR accuracy through leveraging large language models to refine transcriptions. Team \textit{h} explores personalization strategies using speaker vectors. Due to space constraints, Table \ref{table:training_strategies} does not include the methods of all participating teams. Additional explored strategies include multi-task learning with etiology classification, dysarthric speech enhancement, and other advanced techniques.

\begin{table}[t!]
\centering
  \caption{Comparison of model architectures, training strategies, and core techniques of selected teams and the baseline}\label{table:training_strategies}
    \begin{tabular}{l||p{6.8cm}}
    \toprule
     \textbf{T.} & \textbf{Model Descriptions}\\
     \midrule
     
     \textit{\textbf{a}} &  \textbf{Model: Parakeet-tdt-1.1B} \\
     & \textbf{Fine-Tuning}; +\textbf{Audio Segmentation} sentence-level segmentation via forced alignment; +\textbf{Model Merging} using multiple-checkpoint weight averaging. \\
     
     \rowcolor[HTML]{EFEFEF}
     \textit{\textbf{b}} & \textbf{Model: Whisper-large-v3} \\
     \rowcolor[HTML]{EFEFEF}
     & \textbf{Fine-Tuning}; +\textbf{Audio Segmentation} 15-second segmentation via VAD and semi-supervised self-training strategies.\\

     \textit{\textbf{c}} & \textbf{Model: Whisper-large-v2} \\
     & \textbf{Fine-Tuning} using AdaLoRA; +\textbf{Hallucination Reduction} using WhisperX Preprocessing pipeline and Rule-based Postprocessing; +\textbf{Curriculum Learning} with data filtering.\\
     
     \rowcolor[HTML]{EFEFEF}
     \textit{\textbf{d}}  & \textbf{Model: Whisper-large-v3} \\
     \rowcolor[HTML]{EFEFEF}
     & \textbf{Fine-Tuning} using LoRA; +\textbf{Post-ASR Error Correction} by LLM-based Generative Error Collection.\\
     
     \textit{\textbf{e}}  & \textbf{Model: Parakeet-rnnt-0.6B} \\
     & \textbf{Fine-Tuning}; +\textbf{Architecture Comparison} among three architectures, including (Fast)Conformer encoders, CTC/RNNT decoders and ContextNet\cite{han2020contextnet}; \\

     \rowcolor[HTML]{EFEFEF}
     \textit{\textbf{h}}  & \textbf{Model: Whisper-large-v3}\\
     \rowcolor[HTML]{EFEFEF}
     & \textbf{Fine-Tuning} using AdaLoRA; +\textbf{Personalization}: Mapping speaker vectors to latent spaces. \\

     \textbf{*} & \textbf{Model: Whisper-large-v2} w/o fine-tuning \\
    \bottomrule
    \end{tabular}
    \vspace{-1.0em}
\end{table}

\section{Conclusion}
This paper provides an in-depth review of the Interspeech 2025 Speech Accessibility Project (SAP) Challenge, a pioneering effort to advance Automatic Speech Recognition (ASR) for individuals with speech disorders. Using the SAP-240430 dataset, which contains over 400 hours of diverse impaired speech, and evaluating performance through Word Error Rate and Semantic Score, the challenge attracted submissions from 22 teams. Twelve teams outperformed the baseline whisper-large-v2 model, with the leading team achieving the lowest WER of 8.11\%, as well as the highest SemScore of 88.44\%. These achievements establish significant benchmarks, demonstrating the effectiveness of fine-tuning the speech foundation models on the SAP data in improving ASR for impaired speech. They also validate the critical role of large-scale, speaker-independent datasets of impaired speech in advancing ASR performance and generalization for unseen speakers with dysarthria. Advanced strategies that have also been explored by our participants include but are not limited to audio segmentation, model merging, hallucination reduction, curriculum learning, post-ASR error correction, and personalization. Building on the foundation established by this challenge, future research should 
further explore within- and across-group similarities and differences, e.g., among
etiology-based or impairment severity-based populations, to further develop inclusive and effective speech recognition technologies accessible to a broader audience.

\section{Acknowledgements}
The data used for this challenge was made possible by a grant to the University of Illinois from the AI Accessibility Coalition, whose members include Amazon, Apple, Google, Meta,
and Microsoft. This research was supported in part by the Illinois Computes project which is supported by the University of Illinois Urbana-Champaign. The
content is solely the responsibility of the authors and does
not represent the views of the AI Accessibility Coalition, Amazon, Apple, Google, Meta, or Microsoft.

\bibliographystyle{IEEEtran}
\bibliography{mybib}

\begin{thebibliography}{10}
\providecommand{\url}[1]{#1}
\csname url@samestyle\endcsname
\providecommand{\newblock}{\relax}
\providecommand{\bibinfo}[2]{#2}
\providecommand{\BIBentrySTDinterwordspacing}{\spaceskip=0pt\relax}
\providecommand{\BIBentryALTinterwordstretchfactor}{4}
\providecommand{\BIBentryALTinterwordspacing}{\spaceskip=\fontdimen2\font plus
\BIBentryALTinterwordstretchfactor\fontdimen3\font minus \fontdimen4\font\relax}
\providecommand{\BIBforeignlanguage}[2]{{%
\expandafter\ifx\csname l@#1\endcsname\relax
\typeout{** WARNING: IEEEtran.bst: No hyphenation pattern has been}%
\typeout{** loaded for the language `#1'. Using the pattern for}%
\typeout{** the default language instead.}%
\else
\language=\csname l@#1\endcsname
\fi
#2}}
\providecommand{\BIBdecl}{\relax}
\BIBdecl

\bibitem{hasegawa2024community}
M.~Hasegawa-Johnson, X.~Zheng, H.~Kim, C.~Mendes, M.~Dickinson, E.~Hege, C.~Zwilling, M.~M. Channell, L.~Mattie, H.~Hodges \emph{et~al.}, ``Community-supported shared infrastructure in support of speech accessibility,'' \emph{Journal of Speech, Language, and Hearing Research}, vol.~67, no.~11, pp. 4162--4175, 2024.

\bibitem{kim08c_interspeech}
H.~Kim, M.~Hasegawa-Johnson, A.~Perlman, J.~Gunderson, T.~S. Huang, K.~Watkin, and S.~Frame, ``Dysarthric speech database for universal access research,'' in \emph{Interspeech 2008}, 2008, pp. 1741--1744.

\bibitem{yadav2019evalai}
D.~Yadav, R.~Jain, H.~Agrawal, P.~Chattopadhyay, T.~Singh, A.~Jain, S.~B. Singh, S.~Lee, and D.~Batra, ``Evalai: Towards better evaluation systems for ai agents,'' \emph{arXiv preprint arXiv:1902.03570}, 2019.

\bibitem{darley1969differential}
F.~L. Darley, A.~E. Aronson, and J.~R. Brown, ``Differential diagnostic patterns of dysarthria,'' \emph{Journal of Speech and Hearing Research}, vol.~12, pp. 246--269, 1969.

\bibitem{solomon1993speech}
N.~P. Solomon and T.~J. Hixon, ``Speech breathing in parkinson’s disease,'' \emph{Journal of Speech, Language, and Hearing Research}, vol.~36, no.~2, pp. 294--310, 1993.

\bibitem{ramig1995comparison}
L.~O. Ramig, S.~Countryman, L.~L. Thompson, and Y.~Horii, ``Comparison of two forms of intensive speech treatment for parkinson disease,'' \emph{Journal of Speech, Language, and Hearing Research}, vol.~38, no.~6, pp. 1232--1251, 1995.

\bibitem{levelt1983monitoring}
W.~J. Levelt, ``Monitoring and self-repair in speech,'' \emph{Cognition}, vol.~14, no.~1, pp. 41--104, 1983.

\bibitem{howell1991use}
P.~Howell and K.~Young, ``The use of prosody in highlighting alterations in repairs from unrestricted speech,'' \emph{The Quarterly Journal of Experimental Psychology Section A}, vol.~43, no.~3, pp. 733--758, 1991.

\bibitem{zhang21ja_interspeech}
Y.~Zhang, E.~Bakhturina, and B.~Ginsburg, ``{NeMo (Inverse) Text Normalization: From Development to Production},'' in \emph{Proc. Interspeech 2021}, 2021, pp. 4857--4859.

\bibitem{Phukon2024}
B.~Phukon, X.~Zheng, and M.~Hasegawa-Johnson, ``Aligning asr evaluation with human and llm judgments: Intelligibility metrics using phonetic, semantic, and nli approaches,'' \emph{arXiv preprint arXiv:2506.16528}, 2025.

\bibitem{chen2023menli}
Y.~Chen and S.~Eger, ``Menli: Robust evaluation metrics from natural language inference,'' \emph{Transactions of the Association for Computational Linguistics}, vol.~11, pp. 804--825, 2023.

\bibitem{zhang2019bertscore}
T.~Zhang, V.~Kishore, F.~Wu, K.~Q. Weinberger, and Y.~Artzi, ``Bertscore: Evaluating text generation with bert,'' \emph{arXiv preprint arXiv:1904.09675}, 2019.

\bibitem{liu2019roberta}
Y.~Liu, ``Roberta: A robustly optimized bert pretraining approach,'' \emph{arXiv preprint arXiv:1907.11692}, vol. 364, 2019.

\bibitem{radford2023robust}
A.~Radford, J.~W. Kim, T.~Xu, G.~Brockman, C.~McLeavey, and I.~Sutskever, ``Robust speech recognition via large-scale weak supervision,'' in \emph{International conference on machine learning}.\hskip 1em plus 0.5em minus 0.4em\relax PMLR, 2023, pp. 28\,492--28\,518.

\bibitem{han2020contextnet}
\BIBentryALTinterwordspacing
W.~Han, Z.~Zhang, Y.~Zhang, J.~Yu, C.-C. Chiu, J.~Qin, A.~Gulati, R.~Pang, and Y.~Wu, ``Contextnet: Improving convolutional neural networks for automatic speech recognition with global context,'' 2020. [Online]. Available: \url{https://arxiv.org/abs/2005.03191}
\BIBentrySTDinterwordspacing

\end{thebibliography}

\end{document}